\documentclass[runningheads]{llncs}
\usepackage[T1]{fontenc}
\usepackage{graphicx}
\usepackage{tikz}
\usepackage{multirow}
\usepackage{pdflscape}
\usepackage{hyperref}

\usepackage{tikz}
\usepackage{textcomp}
\usepackage[doipre={DOI:~}]{uri}
\usepackage{lipsum}
\newcommand\copyrighttext{%
  \footnotesize This paper has been accepted for publication in the SAMOS2024 conference.
  }
\newcommand{\copyrightnotice}{%
\begin{tikzpicture}[remember picture,overlay]
\node[anchor=south,yshift=10pt] at (current page.south) {\fbox{\parbox{\dimexpr\textwidth-\fboxsep-\fboxrule\relax}{\copyrighttext}}};
\end{tikzpicture}%
}

\begin{document}
\title{Accelerating Depthwise Separable Convolutions on Ultra-Low-Power Devices}
%
%
\author{
Francesco Daghero\inst{1}\orcidID{0000-0001-9595-7216} \and
Alessio Burrello\inst{1}\orcidID{0000-0002-6215-8220}  \and
Massimo Poncino\inst{1}\orcidID{0000-0002-1369-9688} \and
Enrico Macii\inst{1}\orcidID{0000-0001-9046-5618} \and
Daniele Jahier Pagliari\inst{1}\orcidID{0000-0002-2872-7071}
}
\authorrunning{F. Daghero et al.}
\institute{
Politecnico di Torino,
Corso Castelfidardo 39, 10129, Turin, Italy
\email{name.firstsurname@polito.it}\\
}

\maketitle              
\copyrightnotice

\begin{abstract}
Depthwise separable convolutions are a fundamental component in efficient Deep Neural Networks, as they reduce the number of parameters and operations compared to traditional convolutions while maintaining comparable accuracy. However, their low data reuse opportunities make deploying them notoriously difficult.
In this work, we perform an extensive exploration of alternatives to fuse the depthwise and pointwise kernels that constitute the separable convolutional block. Our approach aims to minimize time-consuming memory transfers by combining different data layouts.
When targeting a commercial ultra-low-power device with a three-level memory hierarchy, the GreenWaves GAP8 SoC, we reduce the latency of end-to-end network execution by up to 11.40\%. Furthermore, our kernels reduce activation data movements between L2 and L1 memories by up to 52.97\%.
\end{abstract}

\section{Introduction}
\label{sec:introduction}

A popular trend for AI-based IoT applications consists of performing computations directly on sensing devices, thus avoiding or at least reducing the dependency on a network connection, making applications more secure, energy-efficient, and responsive.
Consequently, much research has been done to devise HW platforms, algorithms, or new operators to enable fast and energy-efficient Deep Neural Networks (DNNs) inference on IoT nodes~\cite{han2015deep,Daghero2020}.
One notable example is represented by \textit{depthwise separable convolutions}, which have progressively become core components of several efficient DNN architectures~\cite{mobilenetV1,s2018mobilenetv2,xception}, thanks to their reduced number of parameters and operations w.r.t. standard convolutions, at the cost of limited or 0 accuracy drops.
On the other hand, these primitives are challenging to accelerate, as they are characterized by a lower data reuse w.r.t. standard convolutions. %

At the same time, edge-oriented hardware architectures are starting to include more complex memory hierarchies, with a smaller but faster L1 memory, where the whole network does not fit, and a bigger but slower L2 memory. These memories are often software-controlled scratchpads to get rid of energy-expensive caches.
Moving data between different memory levels requires a significant amount of energy. Layer fusion~\cite{cudnn,triton} has been introduced to lessen the memory transfers, where two or more layers are merged into one.
This approach avoids moving intermediate tensors back and forth from L1 to L2, thus avoiding time-consuming transfers at the cost of additional overhead in terms of peak memory usage.
However, most fusion optimizations are limited to the simple case of one convolution and an elementwise operator (e.g. Pooling, ReLU, etc)~\cite{layer_fusion}.

\begin{table}[t]
\centering
\footnotesize
\caption{Notation}\label{tab:notation}
\begin{tabular}{cc}
\textbf{Dimension} & \textbf{Abbreviations}   \\\hline
Input (rows,columns,channels)          &  IX/IY/C \\
Output (rows,columns,channels)          &  OX/OY/K \\
Weights (filter height/width, channels in/out)                 & FX/FY/C/K  \\
Padding/Stride & P/S   \\
Fused dimension                & FD  \\ \hline
\end{tabular}
\end{table}
In this work, we propose instead a set of efficient fusion alternatives for depthwise and pointwise convolution sequences aimed at maximizing the data reuse of each primitive while minimizing the data transfers and re-organizations between different memory levels.
Our kernel library is released as open-source at: \url{https://github.com/eml-eda/depthwise-separable-fusion}.
Our main contributions are as follows:
\begin{itemize}
\item We propose and benchmark \textit{six new fused kernels}, each leveraging different data layouts and data processing patterns. On the GreenWaves GAP8 SoC~\cite{flamand2018gap}, considering 36 blocks with different input/output sizes and n. of channels, our most effective solution has a median computational overhead of as low as 5.13\%, when not considering memory transfers. 
\item We expand the open-source AI-compiler DORY~\cite{burrello2020dory} to support fused kernels, adding an engine that chooses which layer to fuse based on graph analysis with pre-defined constraints. Using our kernels as backend and GAP8 as target, we reduce the inference latency of end-to-end Deep Neural Networks (DNNs) execution by up to 11.40\% while reducing activation memory transfers by 27.26\%. When minimizing the number of transfers, we achieve a reduction of up to 52.97\% while reducing the inference latency by 2.64\%.
\end{itemize}

\section{Background \& Related Works}
\label{sec:background_related}

\subsection{Depthwise separable convolutions}\label{subsec:background:dw-pw}
Table~\ref{tab:notation} reports the notation used in the paper for convolution hyperparameters.
Depthwise separable convolutions factorize a standard convolution into two parts: a DepthWise convolution (DW) and a $1\times 1$ convolution called PointWise (PW).
Their equations can be written as follows:
$$
O_{k,h,w}^{DW}=\sum_{i,j}^{F_x, F_y}I_{k,x+i,y+j} \cdot W_{k,i,j} \quad O_{k,x,y}^{PW}= \sum_{c}^{C}I_{c,x,y} \cdot W_{k,c}
$$
where $I$, $W$, and $O$ denote input, weights, and output tensors for a pixel at height $x$, width $y$, and channel $c / k$.
Note that the DW computation is applied to each channel independently.
As an example, a convolution with $C=32$, $K=64$, $OX=OY=56$, $FX=FY=3$, and $S=2$ requires 14.5M multiply-and-accumulate (MAC) operations, while the corresponding depthwise separable block (DW + PW) only 1.8M, with a reduction of 7.9$\times$. Memory is reduced by the same factor, from 18.4 kB to 2.3 kB.

However, DW is characterized by a significantly lower data reuse compared to standard convolutions, requiring careful handling to maximize its efficiency.
Libraries such as CMSIS-NN~\cite{lai2018cmsis} and PULP-NN~\cite{garofalo2020pulp}, implementing state-of-the-art primitives for deep learning, respectively, on ARM and RISC-V SoCs, propose either specific implementations to be selected depending on the kernel dimension, or implementations that require data re-organization before execution.
In particular, PULP-NN converts the input data layout from Height-Width-Channel (HWC), used for all other library layers, to Channel-Height-Width (CHW) for DW, significantly improving the data locality but requiring an additional data re-organization step.
Therefore, the efficiency of such layers remains low: considering the same hyper-parameters of the previous example, the DW layer accounts for only 12.3\% of the total operations of the DW+PW block but requires 59.9\% of the total latency when executed with the PULP-NN library on GAP8~\cite{flamand2018gap}.

\subsection{Depth-First Tiling \& Layer-Fusion}
Data movement between memory levels is a critical problem for deep learning models~\cite{ivanov2021data}, as given modern networks' size, even single layers may not fit in L1~\cite{burrello2020dory}.
A key method to solve this problem is \textit{tiling}, which divides layers into sub-operations, each using only a portion of the inputs/weights and/or computing a portion of the output, but fitting entirely in L1~\cite{burrello2020dory}. Clearly, the overhead of tiling is an increase in the required data transfers. For instance, the same input might be loaded multiple times in different tiles to compute different output channels.

Several works~\cite{mcunetV2,depthfirst} have proposed ways to alleviate this problem, by performing the execution in a ``depth-first'' fashion, that is, processing the same tile over successive layers, rather than completing the execution of all the tiles of a layer before starting with the next one.
The authors of~\cite{mcunetV2} show that, on a MobilenetV2 architecture with an input resolution of 224x224, depth-first tiling reduces the peak memory usage by up to 8$\times$.
However, this comes at the cost of a computational overhead (around 3\%) due to the redundant computations to recalculate intermediate pixels included in the receptive fields of multiple contiguous tiles.
The authors of~\cite{depthfirst} propose a depth-first inference method too, but compared with~\cite{mcunetV2}, instead of square tiles, they use \textit{row} tiles, always splitting the layer on the IY/OY dimension.
Further, they buffer in L1 the intermediate pixels that should be recomputed between adjacent tiles to contain the computational overheads. They reduce memory transfers by up to 5.2x with a small 0.3\% overhead in terms of computations.

In this work, we take inspiration from the row-based inference of~\cite{depthfirst}, combining it with another popular approach to reduce the data movements, i.e., layer fusion.
Most of the currently employed deep learning backends such as CuDNN~\cite{cudnn}, Triton~\cite{triton}, and ONNX runtime~\cite{onnxruntime} support fusion between convolutions and elementwise operators (e.g., ReLU or BatchNorm).
However, more complex fusion patterns, such as the one between DW and PW proposed in this work, are less explored.
To the best of our knowledge, we are the first to release an open-source library for depth-first DW+PW fusion targeted for edge devices.

\subsection{IoT Edge Nodes}
Thanks to their energy efficiency, heterogeneous platforms have become increasingly popular as IoT edge nodes. 
They comprise multiple cores or processing elements covering specific tasks, such as I/O, optimized digital signal processing (DSP), or matrix multiplication acceleration, and they usually employ a multi-level hierarchy of software-controlled memories, with lower levels (L1) featuring fast access but small size, and higher levels being slower but bigger.
Examples of these architectures are already commercialized by NXP~\footnote{\url{https://www.nxp.com/products/processors-and-microcontrollers/arm-microcontrollers/general-purpose-mcus/lpc4300-cortex-m4-m0}}, STM~\footnote{\url{https://www.st.com/en/microcontrollers-microprocessors/stm32h7-series.html}}, and GreenWaves~\cite{flamand2018gap}.
In particular, GreenWaves proposes the GAP8 SoC, which features a single RISC-V core to handle I/O and sensor interfaces and an 8-core RISC-V processor cluster used to speed up DSP.
GAP8 features 64 kB of L1, 512 kB of L2, and an optional external L3 memory, with Direct Memory Access (DMA) controllers to handle data transfers.
\section{Materials \& Methods}
\label{sec:methods}
\subsection{Fused Kernel Design}
\label{subsec:kernels}
This section describes several alternatives to fuse the DW and PW layers, designed to produce different constraints and memory-latency trade-offs.
For all versions, we use $S$ and $FX/FY$ to denote the stride and the filter size of the DW (since all three values are equal to 1 for PW).
The inner kernels that we fuse are written in C language, and are slightly modified versions of the PULP-NN primitives for GAP8~\cite{garofalo2020pulp}. Therefore, all our kernels target multi-core platforms with 8-bit SIMD extensions.
We consider fusing both DW-PW sequences and PW-DW sequences. 
All our fused kernels use intermediate buffers to avoid recomputation since this solution guarantees lower latency~\cite{depthfirst}.
 
\subsubsection{Depthwise-Pointwise}\label{sec:methods_dw_pw}
\begin{figure}[t]
    \centering \includegraphics[width=0.4\columnwidth]{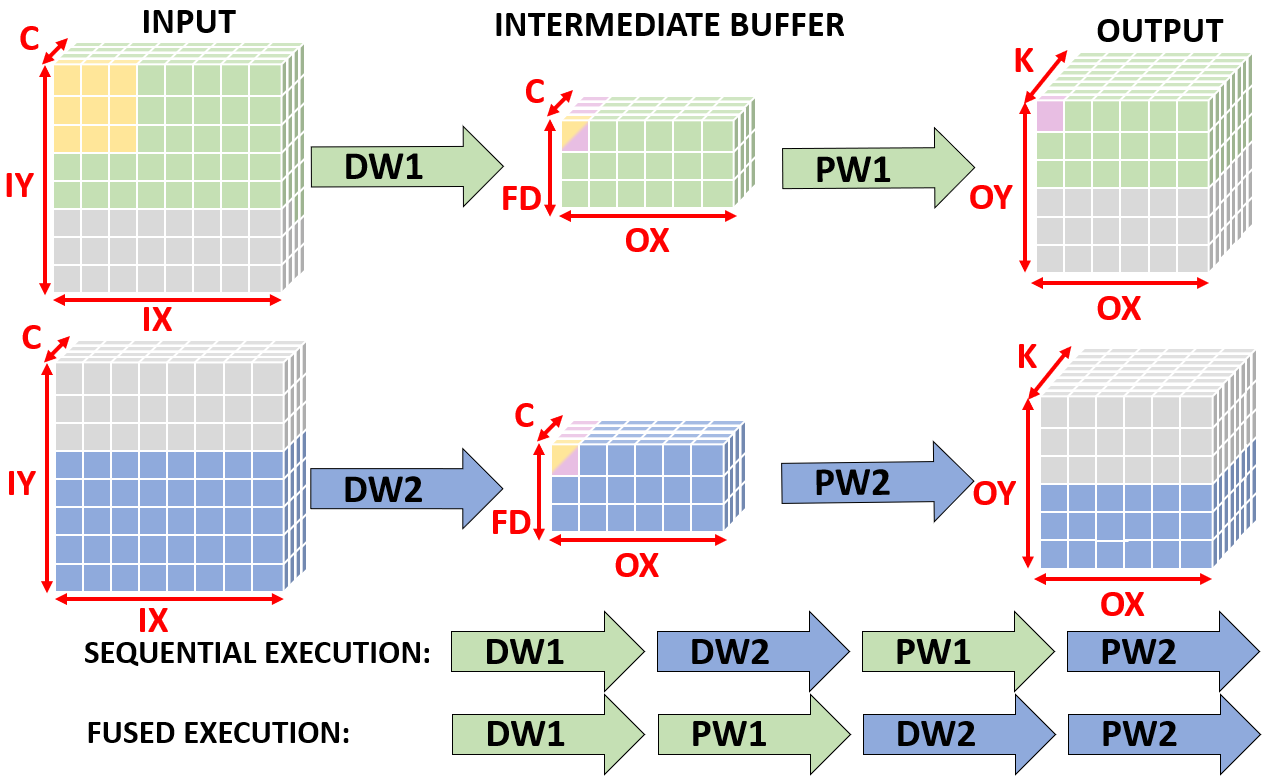} 
    \caption{Execution flow for standard and fused DW-PW sequences, for a layer with $FD=3$, and $IX/IY =8$.}
    \label{fig:dw-pw}
\end{figure}

When fusing the DW-PW sequence, we tile on a per-row basis to limit the overlap between contiguous tiles and, therefore, the additional memory transfers.
Specifically, we divide the layers into blocks of $FD$ rows (we show $FD = 3$ in Fig.~\ref{fig:dw-pw}). In the fused kernel, we first apply the DW operator on each tile, followed by the PW one, as shown by the lowermost set of arrows in the figure. 
Given that the PW layer has a receptive field of 1, this fusion naturally avoids recomputation.
On the other hand, it has two limitations. First, it incurs a memory overhead caused by the intermediate buffer of dimension $C\cdot OX\cdot FD$. Noteworthy, optimal values of $FD$ depend on the parallelization employed by the pointwise kernel. In PULP-NN, each core computes half of an output feature map row, therefore yielding sub-optimal performance if $FD$ is not a multiple of $\frac{N_{cores}}{2}$. As a second limitation, all input channels of the PW must be stored in the intermediate buffer to compute a single output, given that moving partial outputs (in high precision) to L2 would be highly sub-optimal, as explained in~\cite{burrello2020dory}. Consequently, all DW output channels for the same pixel have to be computed before the PW (since $K_{DW}$ = $C_{PW}$), limiting the available tiling options.

\subsubsection{Pointwise-Depthwise}
For the PW-DW sequence, we propose two fused implementations, exploring both a row-wise tiling of the input (as for DW-PW) and a channel-wise one.

\begin{figure}[t]
    \centering
    \includegraphics[width=.80\columnwidth]{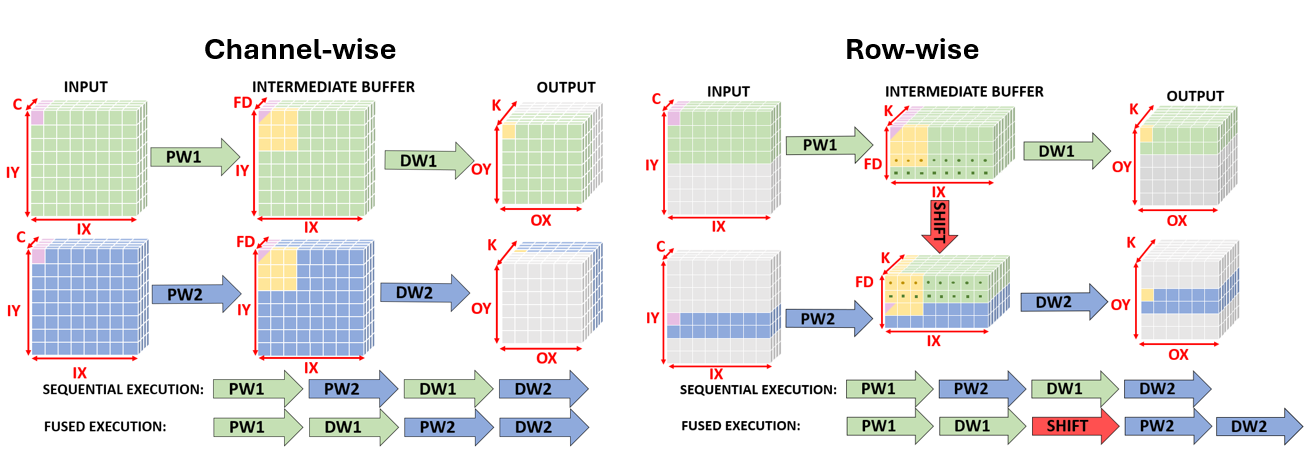} 
    \caption{Execution flow for standard and fused PW-DW sequences for a layer with $FD=4$ and $K=8$ with channel-wise (left) or row-wise (right) tiling.}
    \label{fig:pw-dw}
\end{figure}

\textit{a) Channel-wise execution:}
The left part of Fig.~\ref{fig:pw-dw} provides a high-level overview of the channel-wise tiling, comparing it to an unfused execution. This kernel leverages the independence of the different channels in the DW. Accordingly, we process $FD$ channels per tile using an intermediate buffer of size $FD \cdot IX \cdot IY$. 
Compared to the previous kernel, this version's main advantage is that it enables flexibility in tiling layers also along the channel dimension.
Moreover, recomputation is avoided by having an intermediate buffer that includes all spatial locations of the feature map. 
On the other hand, the intermediate buffer overhead may become substantial for large feature maps, potentially limiting the utility of this kernel. For this solution, optimal values of $FD$ are determined by the DW primitive, which is parallelized over the channels in PULP-NN, leading to performance degradations for $FD$ not multiple of $N_{cores}$.
Another limitation arises from the increased number of input loads required in the PW. In fact, input reuse can no longer be leveraged fully to produce all $K$ output channels, but only a subset of size $FD$.

\textit{b) Row-wise execution:}
We also implement a row-wise fused kernel for the PW-DW sequence, shown in the right part of Fig.~\ref{fig:pw-dw}. Similarly to the DW-PW fusion, and contrarily to the channel-wise PW-DW, this kernel maximizes input data reuse. Furthermore, it maintains the benefits of the channel-wise PW-DW in terms of tiling freedom, given that the channel dimension is not constrained.
However, this fusion requires special care to avoid recomputation: indeed, to execute two adjacent blocks, the DW kernel reuses $FY-1$ rows of its input. 
Accordingly, as depicted in Fig.~\ref{fig:pw-dw}, we shift the intermediate buffer after the completion of each tile, moving the last $FY-1$ rows of the buffer to the beginning. Despite the additional memory movements caused by the shift operation, this solution outperforms one that recomputes PW output rows.

The memory overhead for this kernel is $C\cdot IX\cdot FD$, where a requirement is that $FD\geq FY$, as we need enough rows in the buffer to produce at least one DW output, that is, a number greater or equal to the filter's vertical receptive field.
Above this lower bound, optimal values of $FD$ come from the parallelization of the PW kernel, which as explained in Sec.~\ref{sec:methods_dw_pw} requires $\frac{N_{cores}}{2}$ rows for maximum throughput. As each tile needs $FD - (FY-1)$ new input rows, $FD$ must be a multiple of $\frac{N_{cores}}{2}+FY-1$ to ensure the maximum efficiency of the PW.

\subsection{Data Layout Selection}\label{sec:layout}
To implement our kernels in C, we enhance PULP-NN to have, for both PW and DW, four alternative ``single-layer'' primitives, using all possible combinations of CHW and HWC data layouts for the input and output buffers.
From these, we then create the fused kernels, selecting only combinations that use the same data layout for input and output activations, to ensure that burdensome data re-organization steps are eliminated between concatenated kernels of the same type. 
Since we consider fused kernels for both DW-PW and PW-DW sequences (the latter with two different tiling alternatives), we initially have a pool of 12 kernels. Based on the findings of \cite{garofalo2020pulp}, we then remove layout combinations already identified as suboptimal, such as using CHW for both PW input and DW output in the PW-DW approach, ending up with 6 kernels.
Lastly, we profile these 6 kernels running entirely in L1 on our target platform. Thanks to this profiling, we further restrict the set, considering only the kernels with the lowest latency for each sequence type (2 in total) in end-to-end DNN deployments. The details of this last selection step are reported in Sec.~\ref{sec:kernel_analysis}.

\subsection{End-to-end Network Deployment}\label{sec:end_to_end}
In addition to the kernels, a memory management engine, a tensor allocation engine, and a translator from high-level descriptions (e.g., ONNX) to C kernel function calls are also required to deploy real DNNs on hardware. For this purpose, we leveraged the open-source framework DORY~\cite{burrello2020dory} that comprises all these components, including a tiler that automatically partitions the layers into sub-nodes whose tensors fit entirely in L1 memory.

In particular, we extended DORY to support our kernels with three key innovations: i) a fusion engine, which enables block fusion given an input method (PW-DW or DW-PW) and its constraints; ii) a new tiler for fused kernels, and iii) a post-processing optimizer that, given a profiled execution of every single layer in a network with multiple kernels alternatives, selects the optimal combination to minimize the latency or the number of memory transfers.

The \textbf{fusion engine} determines whether block fusion is possible, given backend constraints, memory constraints, and minimum tile sizes. For example, considering the proposed row-wise PW-DW kernel, we must respect $FD \ge FY$. With a layer with $C = K = 512$, $FX=FY=3$ and $IX=IY=16$,  the smallest tile size is 66.6 kB. If this size exceeds the L1 memory, the engine disables the fusion for the block, and the unfused kernels are employed instead.

The \textbf{tiler for fused layers} is based on the existing tiler in DORY, integrating the aforementioned new constraints from the backends, as well as constraints related to the geometry of fused layers (e.g., the spatial tile size ratio of input, intermediate, and output tiles has to be compatible), and heuristics related to the optimal FD values, as detailed in Sec.~\ref{subsec:kernels}.

Lastly, the \textbf{post-processing optimizer} receives an input flag to select between minimizing the number of cycles or the memory transfers. Below, we detail how the min-latency mode works, but the reasoning is analogous for min-memory.
First, three versions of the complete network, enabling DW-PW fusion, PW-DW fusion, or disabling all fusions, are executed, and the latencies of every layer are collected. For this, we only consider the ``optimal'' combination of fusion type and data layout for each sequence, selected as explained in Sec.~\ref{sec:layout}.
Subsequently, two steps are performed: first, we consider DW-PW and PW-DW separately, comparing the latency of the corresponding fused blocks with the unfused execution of their composing layers, and selecting the lowest latency alternative for each block.
Then, the two resulting \textit{partially-fused} graphs are compared with each other. With an exhaustive search, the fusion option that minimizes latency is selected for each block. Noteworthy, this search has complexity $2^M$, where M is the maximum number of fused blocks (e.g., 13 for a MobileNetV1), which is manageable for edge DNNs.

\section{Experimental Results}
\label{sec:results}
\subsection{Experimental setup}
To benchmark the ideal performance of each kernel in L1, we initially employed the GVSoC virtual platform~\cite{gvsoc}, which allows to simulate a modified version of GAP8 with an increased L1 memory size (1MB in our experiments), thus forcedly avoiding memory transfers between memory levels. End-to-end neural networks are instead deployed on the actual GAP8 hardware, using the modified DORY compiler described in Sec.~\ref{sec:end_to_end} to handle the layers' fusion, tiling, and memory allocations. 
We use the GAPuino board with the GAP8 SoC and an external 8MB flash memory for benchmarking. Memory transfers, cycles, and latency are measured using hardware performance counters at a frequency of 100 MHz.

For assessing the effectiveness of the proposed fused kernels on end-to-end DNN execution, we considered three architectures: i)
MobilenetV1 (MV1)~\cite{mobilenetV1}, whose 29 layers include an initial standard convolution followed by 13 depthwise separable blocks, and final pooling and fully connected layers; ii) MobilenetV2 (MV2)~\cite{s2018mobilenetv2}, with a sequence of 16 \textit{bottleneck blocks} (i.e., PW-DW-PW blocks with residual additions) for a total of 65 layers; iii) DSCNN~\cite{zhang2017hello}, composed of 4 DW-PW blocks and 9 total layers. For both MV1 and MV2, we use a width multiplier of 0.25 and benchmark both on 224$\times$224 and 128$\times$128 inputs, as in the original papers. 
Furthermore, we also consider the MV1 variant proposed in the MLPerf Tiny Suite, which uses a 96$\times$96 input for a visual wakeword task.
\begin{table*}[ht]
\centering
\footnotesize
\caption{Mobilenets-like architectures deployed on the GAP8 platforms. Abbreviations: B. Baseline, LL/LMT Lowest Latency/Memory Transfer, L Latency, WM Weights Memory, AMT Activation Memory Transfers}\label{tab:results-dory}
\begin{tabular}{clcclcc}
\multicolumn{1}{l}{\textbf{Model}} & \textbf{Method}                 & \multicolumn{1}{l}{\textbf{MACs}} & \multicolumn{1}{l}{\textbf{Tot.Cycles [\#]}} & \textbf{L {[}ms{]}} & \multicolumn{1}{l}{\textbf{WM {[}B{]}}} & \multicolumn{1}{c}{\textbf{AMT [\#]}} \\\hline\hline
\multirow{3}{*}{MV1-224}                & B                        & \multirow{3}{*}{41.0M}         & 10.0M                                & 100.6                    & \multirow{3}{*}{463.6k}                            & 2.9M\\
                                          & LL &                                   & 8.9M {[}-11.16\%{]}                  & 89.4                      &                                                   & 2.75M {[}-5.39\%{]}                           \\
                                          & LMT &                                   & 11.4M {[}+12.87\%{]}                  & 113.6                    &                                                   & 1.64M {[}-43.56\%{]}                          \\\hline
\multirow{3}{*}{MV1-128}                & B                        & \multirow{3}{*}{13.57M}         & 3.08M                                 & 30.83                     & \multirow{3}{*}{463.6k}                            & 879.0k                                          \\
                                          & LL &                                   & 2.73M {[}-11.40\%{]}                  & 27.31                     &                                                   & 639.4k {[}-27.26\%{]}                           \\
                                          & LMT &                                   & 3.07M {[}-0.23\%{]}                   & 30.76                     &                                                   & 491.9k {[}-47.74\%{]}                           \\\hline
\multirow{3}{*}{MV1-96}                      & B                        & \multirow{3}{*}{7.49M}          & 2.20M                                 & 22.04                     & \multirow{3}{*}{208.1k}                          & 497.03k                                          \\
                                          & LL &                                   & 2.02M {[}-8.43\%{]}                   & 20.18                     &                                                   & 343.81k {[}-30.83\%{]}                           \\
                                          & LMT &                                   & 2.03M {[}-8.03\%{]}                   & 20.27                     &                                                   & 270.09k {[}-45.66\%{]}                           \\ \hline\hline
\multirow{3}{*}{MV2-224}                & B                        & \multirow{3}{*}{37.20M}         & 14.89M                                & 148.88                    & \multirow{3}{*}{1.51M}                         & 5.83M                                         \\
                                          & LL &                                   & 13.27M {[}-10.85\%{]}                 & 132.72                    &                                                   & 3.51M {[}-39.84\%{]}                           \\
                                          & LMT &                                   & 13.43M {[}-9.73\%{]}                  & 134.38                     &                                                   & 3.28M {[}-43.80\%{]}                           \\\hline
\multirow{3}{*}{MV2-128}                & B                        & \multirow{3}{*}{13.01M}         & 6.76M                                 & 67.62                     & \multirow{3}{*}{1.51M}                         & 2.47M                                         \\
                                          & LL &                                   & 6.33M {[}-6.43\%{]}                   & 63.27                     &                                                   & 1.72M {[}-30.39\%{]}                           \\
                                          & LMT &                                   & 6.58M {[}-2.64\%{]}                   & 65.83                     &                                                   & 1.16M {[}-52.97\%{]}                           \\\hline\hline

\multirow{3}{*}{DSCNN}                    & B                        & \multirow{3}{*}{2.66M}          & 686.25k                                  & 6.86                      & \multirow{3}{*}{22.01k}                           & 144.67k                                          \\
                                          & LL       &                                   & 685.79k {[}-0.07\%{]}                    & 6.86                      &                                                   & 128.67k {[}-11.06\%{]}                           \\
                                          & LMT &                                   & 761.42k {[}+10.95\%{]}                    & 7.61                      &                                                   & 80.67k {[}-44.24\%{]} \\   \hline\hline
\end{tabular}
\end{table*}
\subsection{Kernel analysis}\label{sec:kernel_analysis}

First, we benchmark the performance of individual kernels with all tensors stored in L1 memory and, therefore, neglecting the overhead of memory transfers and memory re-organization steps between layers. With an infinite L1, fused and unfused latencies should ideally be identical. Therefore, this experiment allows us to analyze non-idealities related to the implementation of each kernel, e.g., due to strided memory accesses, control flow overheads, etc.

\begin{figure}[t]
    \centering
    \includegraphics[width=.92\columnwidth]{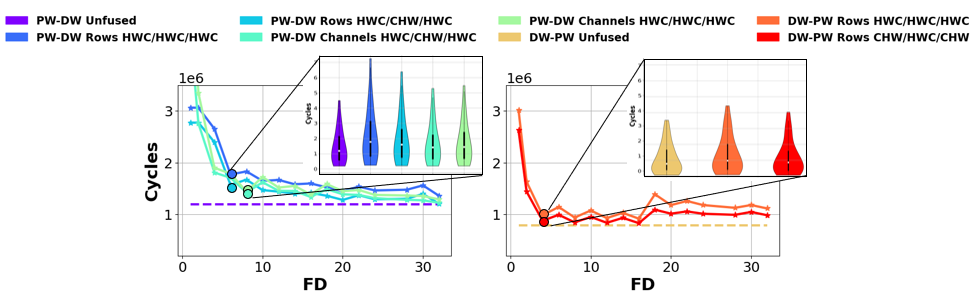} 
    \caption{Median execution cycles in L1 when changing the dimension of the intermediate buffer ($FD$). }
    \label{fig:results-l1-cycles-per-ib}
\end{figure}

Figure~\ref{fig:results-l1-cycles-per-ib} reports the median cycles over 36 DW-PW / PW-DW blocks with different geometries for our six fused kernel variants and for the two unfused baselines, as a function of the $FD$ value. 
We consider $IX=IY \in [32,64,128]$, $S \in [1,2]$, $C \in [32,64,128]$ and $K \in [C,2C]$. 
Violin plots show the distribution of the performance over the 36 geometries at the minimum $FD$ that allows the full exploitation of the multi-core parallelization on GAP8.

The legend uses a compact string to identify each kernel. For instance, \textit{PW-DW Channels HWC/CHW/HWC} denote PW-DW fused kernels with tiling on the channels, using HWC as input/output data layout and CHW for the intermediate buffer.
Overall, for PW-DW fusion, the top-performing kernel is \textit{Rows HWC/CHW/HWC}, with a cycles' overhead of 7.15\% at $FD=20$. This is because the overhead of the shift operation becomes negligible compared to the execution of the whole kernel with large FD values.
On the other hand, large values of $FD$ are often unfeasible when considering a realistic L1 size, as the memory overhead of the intermediate buffer would reduce the available L1 memory for input, output, and weights tensors. 
Restricting to $FD<10$, the \textit{Channels HWC/CHW/HWC} at $FD=8$ achieves the lowest overhead of 19.27\%. Note that 8 is the smallest value of FD that leads to fully utilizing the 8 cores of GAP8.
This kernel is superior to the version that uses an HWC intermediate data layout because a CHW input to the DW avoids strided loads, which would incur a higher latency overhead than strided stores of the PW output.

Concerning the DW-PW fusion, the top performing kernel across all FD values is \textit{Rows CHW/HWC/CHW}. The minimal overhead of 5.13\% is attained for $FD=16$.
Similarly to the PW-DW fusion, the optimal layout combination favors strided stores of the output over strided loads of the input. It's worth noting that this outcome aligns with the inherent structure of the PULP-NN functions. In fact, given the kernels' output stationary dataflow, each output is written only once, whereas inputs are reused for multiple convolution channels and/or multiple output spatial locations. Consequently, overheads associated with loading input data are more detrimental than those related to storing outputs. We also notice that at $FD=4$ the kernel achieves an overhead of 11\%, just 4\% higher compared to the one obtained at $FD=8$, while halving the memory overhead.

As a result of these analyses, we employ the \textit{Channels} \textit{HWC/CHW/HWC} fusion for PW-DW blocks and the \textit{Rows CHW/HWC/CHW} one for the DW-PW blocks, to minimize the optimization complexity when deploying end-to-end neural networks, as anticipated in Sec.~\ref{sec:end_to_end}.
For both kernels, FD is set to the minimal value that ensures utilization of all cores, i.e., $FD=8$ and $FD=4$.%

\subsection{End-to-End Network Deployment}

Table~\ref{tab:results-dory} reports the performance of end-to-end networks with three different configurations: the un-fused Baseline, the fused combination that minimizes the latency, and the one that minimizes the number of memory transfers (which are a good proxy for energy consumption).
We achieve the largest latency reduction on MV1-128 (11.40\%), simultaneously reducing the memory transfers by 27.26\%.
Despite being characterized by more memory transfers (2.9M vs. 879k), MV1-224, achieves a slightly lower latency improvement of 11.16\%, with a reduction in the number of transfers of only 5.39\%. Interestingly, we also notice that by setting our post-processing optimizer to min-memory mode, while we can cut the number of transfers by roughly the same amount on MV1-128 and MV1-224, the former has a latency comparable to the baseline, while the latter incurs a 12.87\% latency increase.
While counter-intuitive, this result stems from the higher resolution of the input feature map, which forces the layers' tiles to be bigger on spatial dimensions. In fact, each fused tile is constrained to include either a whole number of rows or a whole number of complete feature maps, depending on the chosen kernel. This in turn, reduces the tile size on the output channels and, consequently, the input data reuse.
The MV1-96 variant achieves a slightly lower latency reduction of 6.43\%, with a memory transfer reduction of 45.66\%.
This seems to contradict the previous observation but is a consequence of a significantly reduced number of transfers compared to the MV1-128 network (497.03k vs. 879k), which in turn diminishes their impact on the overall network latency, thereby reducing the possibility of optimization through layer fusion.

Concerning the MV2 architecture, the MV2-224 variant achieves the best latency savings w.r.t. the baseline (10.85\%).
By fusing the kernels, both this network and the min. transfer one avoid expensive activation transfers from L3, thus reducing latency significantly.
Nonetheless, MV2-128 also achieves a relevant cycles' reduction of 6.43\% and a memory transfer saving of 52.97\%.

Finally, DSCNN achieves the lowest gains in terms of latency, 0.07\%. This is due to the large size of the intermediate features maps, caused by a large number of channels, which creates problems both with the DW-PW intermediate channels' constraint and with the PW-DW input data reuse. When the goal is minimizing memory transfers, we achieve a reduction of 44.24\%. 

To exemplify the energy gains from memory transfer reductions, we refer to the energy benchmarking of ~\cite{vega}, on a chip with the same architecture of GAP8, albeit on a different technology node. Using those numbers to estimate the energy for computing, L2-L1 transfers, and L3-L2 transfers, and using our results on MV2-224, we obtain an estimated energy reduction of 90.94\%. Savings are high in this case as the original network involves expensive L3 accesses, but this depends on the target's memory hierarchy, which is orthogonal to our approach.

\begin{figure}[t]
    \centering
    \includegraphics[width=0.8\columnwidth]{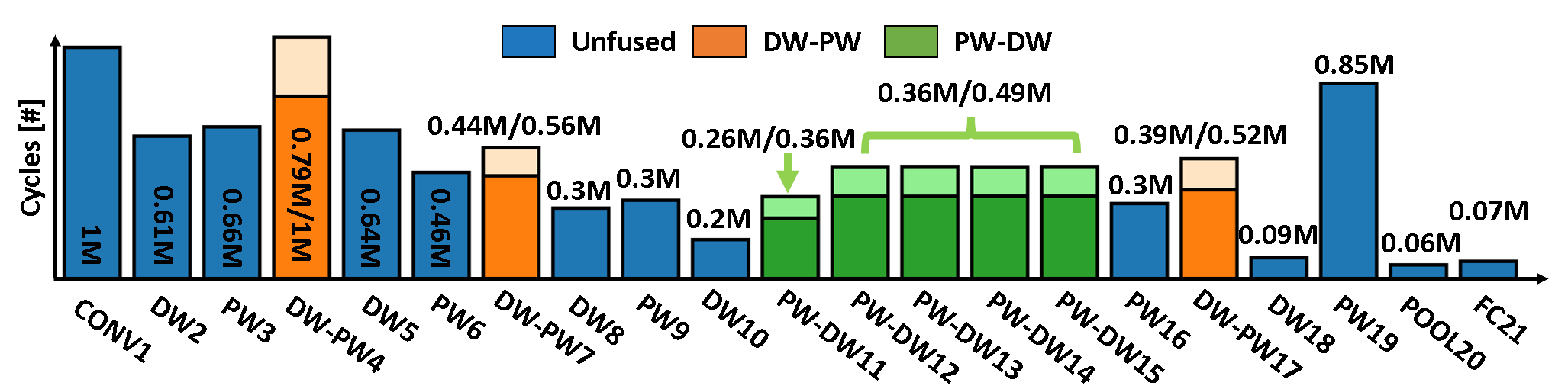} 
    \caption{Layer-by-layer execution cycles and memory transfers of the MV1-224 "Lowest Latency". The lighter colors show the improvement compared to the unfused execution.}
    \label{fig:mv1-cycles}
\end{figure}

\textit{a) End-to-End Network Analysis: }
Figure~\ref{fig:mv1-cycles} depicts the fusion strategies selected by DORY for the ``Lowest Latency'' MV1-224 deployment. 
A first important consideration is that the engine combines unfused kernels and both types of fused sequences, demonstrating that all three strategies are relevant and that the choice between them strongly depends on the layer's geometric parameters.

In the early layers of the network, the (row-wise) DW-PW fusion is advantageous, as it allows the computation of all output channels without re-loading the input multiple times. Note that, as shown in Sec. \ref{sec:kernel_analysis}, this fusion is also the one causing the lowest computational overhead. PW-DW fusion becomes more favorable in the network's core, where the number of channels increases. Indeed, this kernel effectively addresses the tiling issue stemming from the DW-PW kernel's constraint of storing all intermediate channels in L1, which negatively impacts performance by leading to small spatial tiles. Lastly, in the final stages of the network, DW-PW fusion is again employed once. This decision by the optimizer is prompted by a 4$\times$ reduction in spatial feature map size, rendering spatial tiling unnecessary, and therefore making the DW-PW the best alternative again.

\section{Conclusions}
\label{sec:conclusions}
Layer fusion is a well-known strategy to reduce the number of data transfers in DNNs. However, common DNN frameworks apply it only between convolutions and elementwise operators. This work proposes several efficient fusion strategies for DW-PW and PW-DW blocks.
After extensive benchmarking, we integrate the best-performing fused kernels into an open-source framework for end-to-end network deployment.
With experiments on 6 different networks and on the GAP8 SoC, we achieve a latency reduction of up to 11.40\% and a maximum reduction of memory transfers of up to 52.97\%.
\begin{credits}
\subsubsection{\ackname} This work has received funding from the Key Digital Technologies Joint Undertaking (KDT-JU) under grant agreement No 101095947. The JU receives support from the European Union’s Horizon Europe research and innovation program.
\subsubsection{\discintname} The authors have no competing interests to declare that are relevant to the content of this article.
\end{credits}
%
%

%
%
%
\bibliographystyle{splncs04}

\end{document}